%% file: main_arxiv.tex
\documentclass[runningheads]{llncs}
\usepackage{graphicx}
\usepackage{amsmath,amssymb} 
\usepackage{color}

\usepackage{amsmath,amssymb} 
\usepackage{booktabs}
\usepackage{cleveref}
\usepackage{color}
\usepackage{multirow}
\usepackage{siunitx,etoolbox}
\usepackage{pgfplots}
\usepackage{graphics}

\robustify\bfseries
\usepackage{subcaption}
\usepackage{makeidx}
\usepackage{graphicx}
\usepackage{amsmath,amssymb} 
\usepackage{booktabs}
\usepackage{cleveref}
\usepackage{color}
\usepackage{multirow}
\usepackage{siunitx,etoolbox}
\usepackage{pgfplots}
\usepackage{graphics}

\robustify\bfseries
\usepackage{subcaption}
\usepackage[colorinlistoftodos,prependcaption,textsize=tiny,textwidth=20mm]{todonotes}

\begin{document}
\title{Counting the uncountable: Deep semantic density estimation from space} 


\titlerunning{Counting the uncountable: Deep semantic density estimation from space}
%
\author{Andres C. Rodriguez
	\and
Jan D. Wegner } 
%
\index{Rodriguez, Andres C}
\index{Wegner, Jan D}
\authorrunning{A.C. Rodriguez and J.D. Wegner}
%
\institute{ETH Zurich, Stefano-franscini-platz 5 8093 Zurich, Switzerland 
\email{\{andres.rodriguez,jan.wegner\}@geod.baug.ethz.ch}
}

\maketitle              
\begin{center}
\textit{Accepted at GCPR 2018}
\end{center}

\begin{abstract}
We propose a new method to count objects of specific categories that are significantly smaller than the ground sampling distance of a satellite image. This task is hard due to the cluttered nature of scenes where different object categories occur. Target objects can be partially occluded, vary in appearance within the same class and look alike to different categories. Since traditional object detection is infeasible due to the small size of objects with respect to the pixel size, we cast object counting as a density estimation problem. To distinguish objects of different classes, our approach combines density estimation with semantic segmentation in an end-to-end learnable convolutional neural network (CNN).  Experiments show that deep semantic density estimation can robustly count objects of various classes in cluttered scenes. Experiments also suggest that we need specific CNN architectures in remote sensing instead of blindly applying existing ones from computer vision.

\end{abstract}

\input{introduction}

\input{relatedwork}

\input{methods}

\input{experiments}

\input{conclusion}

\subsection*{Acknowledments}
This project is funded by Barry Callebaut Sourcing AG as a part of a Research Project Agreement with ETH Zurich.
\bibliographystyle{splncs04}
	\bibliography{egbib}
%
%
%
%
\end{document}

%% file: introduction.tex
	\section{Introduction}
	\label{sec:introduction}

We propose deep semantic density estimation for objects of sub-pixel size in satellite images. Satellite image interpretation is a challenging but very relevant research topic in remote sensing, ecology, agriculture, economics, and cartographic mapping. Since the launch of the Sentinel-2 satellite configuration of the European Space Agency (ESA) in 2015, anyone can download multi-spectral images of up to 10 meter ground sampling distance (GSD) for free. At the same time, Sentinel-2 offers high revisit frequency delivering an image of the same spot on earth roughly every 5 days. However, for applications that need more high-resolution evidence like detecting and counting objects (e.g, supply chain management, financial industry), spatial resolution is too poor to apply traditional object detectors. In this work, we thus propose to circumnavigate explicit object detection by turning the counting problem into a density estimation task. Furthermore, we add semantic segmentation to be able to count objects of very specific object categories embedded in cluttered background. We integrate semantic segmentation and density estimation into one concise, end-to-end learnable deep convolutional neural network (CNN) to count objects of $1/3$ the size of the GSD.    
\begin{figure}[!t]
	\centering
	\begin{subfigure}[b]{1\textwidth}
		\includegraphics[width=0.5\textwidth, angle = 90, ,trim=16cm 10cm 0cm 0cm, clip=true]{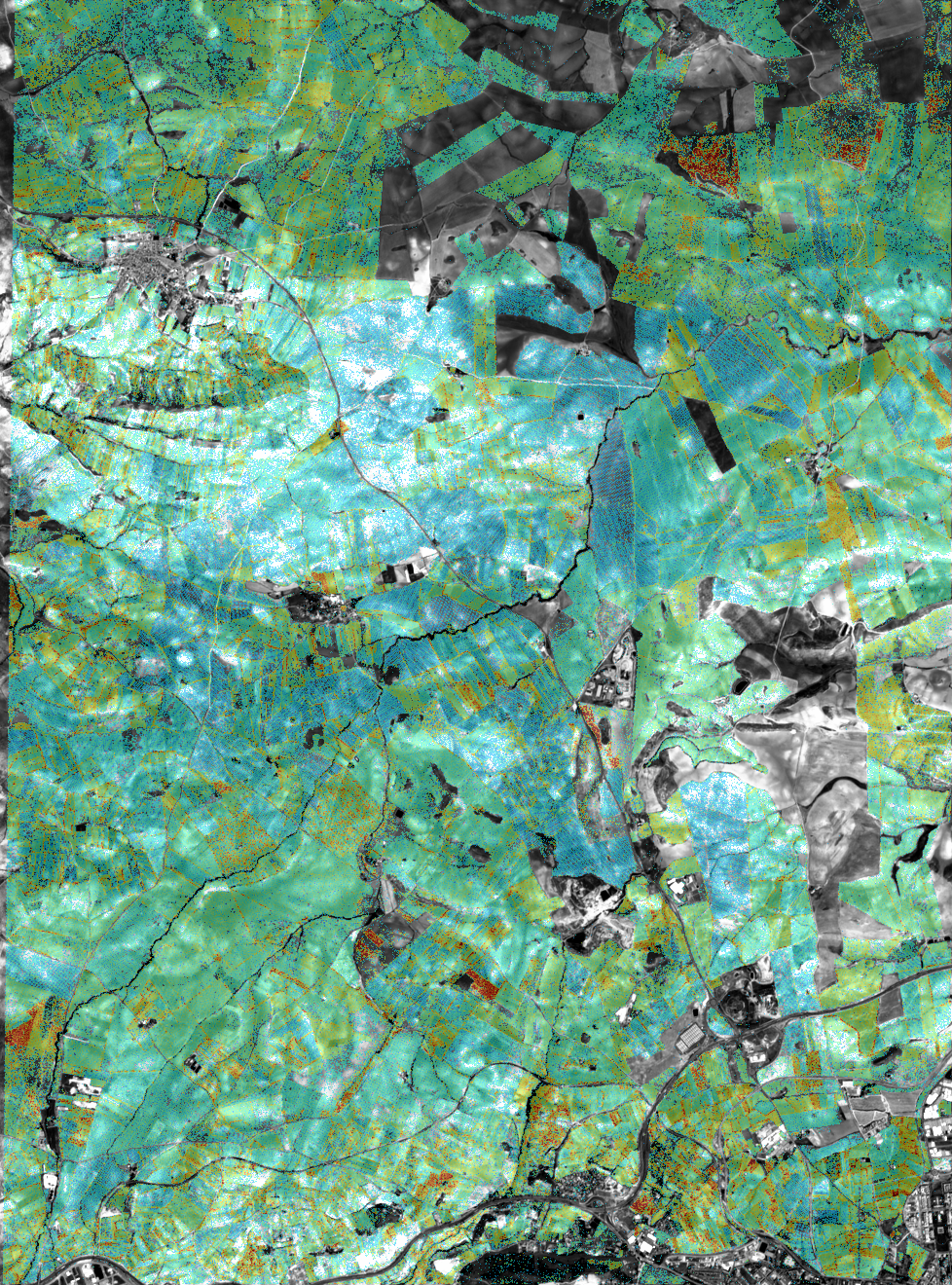}
	\end{subfigure}
	\begin{subfigure}[b]{0.85\textwidth}
		\includegraphics[width=1\linewidth]{{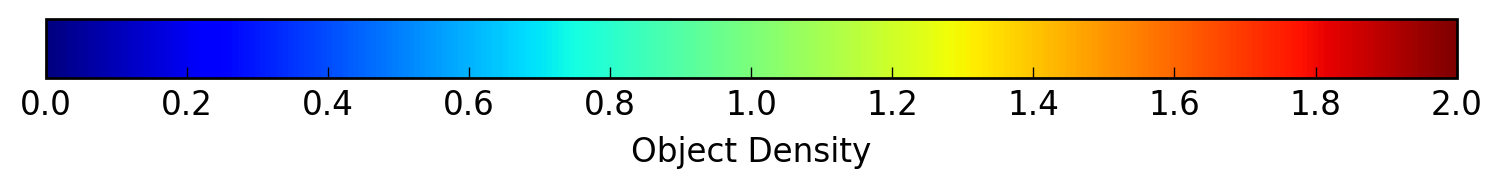}}
	\end{subfigure}	
	
	\caption{Density of Olive oil trees in Ja\'en, Spain, overlaid to a greyscale version of the aerial image. Densities below 0.5 were trimmed for visualization.}
	\label{fig:olives_RGBGT_Dens}
\end{figure}
Semantic Segmentation is a standard task in computer vision and has seen significant performance gains since the comeback of deep learning. The major goal is predicting a class label per pixel over an entire image. A rich set of benchmark challenges and datasets like Cityscapes \cite{Cordts2016Cityscapes} and Pascal VOC 2012 \cite{pascalVoc2012} help making rapid progress in this field while continuously reporting current state-of-the-art in online rankings. Typical objects classes of interest are buildings, persons and vehicles that are (i) clearly visible in the image, (ii) large in size (usually several hundreds of pixels) (iii) and can be distinguished from background and further classes primarily relying on shape and RGB texture. Much of this can be transferred with minor adaptions to overhead imagery of $<2m$ GSD acquired by drones, aerial sensors, and very high-resolution spaceborne platforms \cite{marmanis2016semantic,mattyus2017deeproadmapper,postadjian2017investigating,KuoDeepAggregation2018}\footnote{see the ISPRS semantic labeling benchmark for an overview\url{http://www2.isprs.org/commissions/comm3/wg4/results.html}}. 
In contrast, objects like cars, trees, and buildings constitute a single pixel or less in remote sensing imagery of Sentinel-2 (10m GSD) and Landsat (40m GSD). While this resolution is sufficient for semantic segmentation of large, homogeneous regions like crops \cite{Russwurm_2017_CVPR_Workshops,Pryzant_2017_CVPR_Workshops}, counting individual object instances in cluttered background becomes hard. Here, high spectral resolution comes to the rescue. Deep learning techniques can greatly benefit from high spectral resolution that conveys object information invisible to human sight. It can learn complex relations between spectral bands to identify object-specific spectral signatures that support pixel-accurate semantic segmentation.  
%

Our workflow goes as follows: We manually annotate a small sample of the object class of interest in Google Earth aerial images. Large-scale groundtruth is obtained by training Faster R-CNN~\cite{RenFasterRCNN2015} and then predicting for thousands of object instances. Object detections are manually cleansed and turned into smooth density estimates by filtering with a smoothing kernel. An example for groundtruth of olive tree density is given in Fig.\ref{fig:olives_RGBGT_Dens}, where we plot the density obtained from high-resolution aerial images. This reference is used to train, validate, and test our deep semantic density estimation model using Sentinel-2 images. It turns out that deep semantic density estimation can robustly count specific object instances of size well below the GSD by making use of both, spatial texture and spectral signature. We demonstrate that our method robustly scales to $>200 km^{2}$ counting $>1.6$ Mio. object instances. 

Our main contributions are: (1) We introduce end-to-end learnable deep semantic density estimation for objects of sub-pixel size, (2) we provide a simplified network architecture that takes advantage of low spatial but high spectral image resolution, (3) we show that standard network architectures from computer vision are inappropriate for this task, and (4) we provide a new, large-scale dataset for object counting in satellite images of moderate resolution. 


%% file: relatedwork.tex
\section{Related Work}
	\label{sec:related_work}
There is ample literature for semantic segmentation and object detection in both, computer vision and remote sensing. A full review is beyond the scope of this paper and the reader is referred to top-scoring submissions of benchmarks like Cityscapes \cite{Cordts2016Cityscapes}, MSCOCO~\cite{Lin2014}, and the ISPRS semantic labeling challenge.

\paragraph{Semantic Segmentation} has seen significant progress since the invention of fully convolutional neural networks (FCN)~\cite{long2015fully}. Various improvements to the original architecture have been proposed. For example, more complex Encoder-Decoder models learn features in a lower-dimensional representation and then upscale to the original resolution using deconvolutional layers, making use of middle layer representations or bilinear interpolation \cite{RonnebergerUnet2015,BadrinarayananSegNet2015,LinRefineNet2016,PohlenFullResidual2016}. Further ideas include atrous convolutions to prevent lowering the resolution of the learned features and keeping a large receptive field \cite{DeeplabChen_2016,Rethinking_chen2017,ChenEncDecAtrous2018}. Other approaches are based on pyramid pooling where features are learned in different resolutions and then merged in different ways to deal with objects of different sizes and scales \cite{Rethinking_chen2017,ChenEncDecAtrous2018,LiuParseNet2015}.

\paragraph{Object Detection} is related to semantic segmentation. Instead of assigning one class label per pixel, its objective is detecting all instances of a specified object class in an image. This translates to automatically drawing bounding boxes around objects but leaving background unlabeled. One of the most widely used deep learning detectors is Faster R-CNN~\cite{RenFasterRCNN2015}, which extracts scale-invariant features for object proposals. Other approaches compute features for different scales and extract the most relevant scale-features~\cite{LiuSSD2015}. A more sophisticated variant of object detection is instance segmentation, which adds detailed per-pixel boundaries per object. Today's quasi gold standard is Mask R-CNN~\cite{HeMaskRCNN2017} that builds on Faster R-CNN and adds a dense feature map to predict instance boundaries.

\paragraph{Density Estimation} is about predicting the local distribution of objects without explicitly detecting them. It is a good option in cases where actual objects cannot be clearly identified due to low spatial image resolution, occlusions etc. One application scenario in computer vision is crowd analysis, where ground level images or image sequences from surveillance cameras are used to estimate the number of people~\cite{meynberg2016detection,ShangCrowdCounting2016}. An interesting strategy to ease the learning procedure are ranking operations~\cite{LiuLeveragingRank2018} for density estimation. See~\cite{SindagiSurveyDensity2017} for a complete review of crowd counting and density estimation. 
In remote sensing, density estimation has been used for a variety of applications. 
\cite{doupe2016equitable,robinson2017deep} use deep learning to estimate population density using satellite imagery across large regions. Similarly,~\cite{zhang2017} estimate densities of buildings in cities with support vector regression and a rich set of texture features. More closely related to our work,~\cite{joshi2006} estimate forest canopy density from Landsat images, whereas~\cite{mutanga2012} estimate biomass density in wetlands from Worldview-2 imagery. These methods inspire our research because they demonstrate that satellite imagery can be used for density estimates of objects with sizes significantly below the GSD. However, all methods have in common that they compute densities of objects that are densely populating the respective image without (much) clutter. In contrast, we propose fine-grained, species-specific density estimates in scenes where the target class could be one out of many.
To the best of our knowledge, we are the first to couple fine-grained semantic segmentation with density estimation for cluttered scenes in satellite images. Moreover, we formulate our method as an end-to-end learnable CNN that learns all features for semantics and density estimation simultaneously from the data. 


%% file: methods.tex
\section{Methods}
	\label{sec:methods}

In this section we describe the technical approach. One problem with basically unrecognizable objects in satellite images is manual annotation of reference data. It is impossible to manually segment individual trees or cars in the Sentinel-2 images of 10 meters GSD. We thus resort to Google Maps overhead images of much higher resolution for groundtruth labeling. 

\subsubsection{Ground Truth}

We apply the Faster R-CNN Object detector~\cite{RenFasterRCNN2015} to very high-resolution (1$m$ GSD) Google Maps images to identify and count reference objects. The detector is tuned to achieve high recall to then manually remove false positive predictions. This allows us to obtain a highly detailed count per area that is used as ground truth to test our model on lower resolution satellite imagery. To down-sample our ground truth to a 10$m$ resolution image, we apply a Gaussian kernel with a $\sigma = K / \pi$, where $K$ is the down-scale ratio between the high resolution Google Maps images and the lower resolution satellite image (we chose $K = 10$). We compute the mean over a window of $K\times K$ pixels to obtain the sub-sampled ground truth. The semantic segmentation label is obtained by thresholding densities with a score above $0.5$ as object class, and as background otherwise. Note that we use only cloud-free Sentinel-2 images that are closest to the sensing date of the high resolution imagery in Google Maps. Most Sentinel-2 bands have larger GSD than 10m and we thus need to resample them. We upscale the 20m GSD bands (B5, B6, B7, B8, B10, B11) to 10m using bilinear interpolation. 60m bands were not used in our experiments due to their low resolution and the low information they convey for vegetation datasets.


\subsubsection{Semantic Segmentation and Density Estimation}


Although semantic segmentation and density estimation are different tasks, they are related and can benefit each other. For mutual, fruitful cooperation of both tasks, we set up a joint training procedure such that most features can be used for either of the goals. Adding the semantic segmentation to density estimation also allows to compute classification metrics from standard semantic segmentation literature. Our approach consists of using ResNet blocks, which contain mainly convolutional layers, to obtain features from the input image. Then, for each task, we add an independent convolutional layer at the end of each architecture. 
For all training experiments, our Loss is defined as: 
\begin{equation}
Loss =  \underbrace{CrossEntropy(y_{label}, \hat{y}_{label}) }_{L_{semantic}} +  \underbrace{ (y_{density} - \hat{y}_{density})^2 }_{L_{density}}
\end{equation}


where $\hat{y}$ is the predicted estimate for each task.
We test four different network architectures in our experiments to verify which provides the best basis for our approach.  We begin with the DeepLab V2 (\emph{DL2}) architecture~\cite{DeeplabChen_2016}, which uses atrous convolutions to obtain filters that have a larger receptive field. These convolutions, inspired from signal processing ``\`a-trous algorithm" \cite{MallatWavelet2008}, have filters that are dilated with zeros to increase the spatial view of a filter without increasing the number of parameters. 
Moreover, we experiment with the updated version DeepLab V3 (\emph{DL3}) \cite{Rethinking_chen2017}, where pyramid pooling with several atrous convolution filters is implemented. Note that this method is a top performing architecture for semantic segmentation on Cityscapes and PASCAL VOC 2012 datasets, and can therefore be viewed as state-of-the-art in semantic image segmentation. Our third architecture (\emph{Ours}) consists of a simplified 6 Layers ResNet~\cite{he2016} where our fourth variant (\emph{Ours Atrous}) adds a last layer of atrous convolutions.

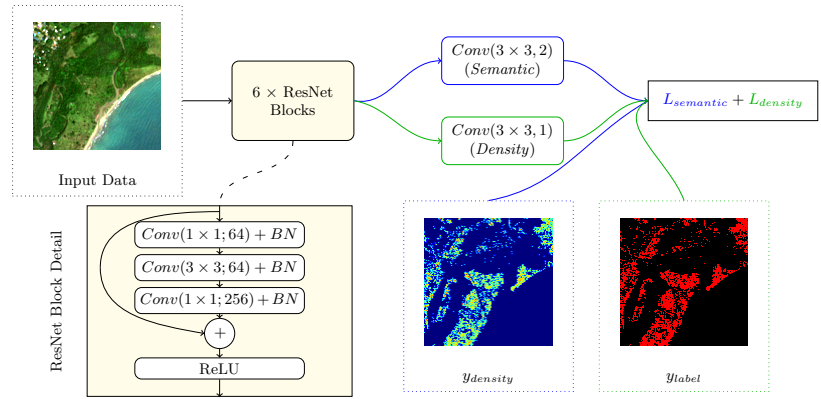
\begin{figure*}[t!]
	\centering
	\scalebox{0.65}{\input{figures/fig_architecture}}
	\caption{\emph{Ours} architecture: $Conv(n\times n,M)$ convolution with $M$ filters of size $n\times n$. $BN$: batch normalization. Input data is first passed through a $Conv(3\times 3,256) + BN$ layer before being fed to the first ResNetBlock. Blue and green colors indicate semantic and density connections, respectively. For \emph{Ours Atrous}, all convolutions of the last block before the loss are changed to atrous convolutions.}
	\label{fig:feat_extractors}
\end{figure*}

For both simplified ResNet architectures (\emph{Ours} and \emph{Ours Atrous}) we change all striding operations to 1 and use 6 consecutive ResNet Blocks (Fig.~\ref{fig:feat_extractors}). Our method moves away from lower dimensional representations but instead keeps details. 6 layers showed a good performance where deeper networks only had marginal to insignificant improvements at a much higher computational cost. The original architecture was designed for typical computer vision images with few, relatively large objects ($>100$ pixels per object) per image. In our case, thousands of object instances of sub-pixel size are present in the image. We want to retain as many details as possible and therefore set stride to 1. We compensate the potential increase in parameters by reducing the size of the receptive field. In fact, we can tolerate learning contextual knowledge within smaller neighborhoods because it is less important in our case.

%% file: figures/fig_architecture.tex
\pgfdeclarelayer{background}
\pgfdeclarelayer{foreground}
\pgfsetlayers{background,main,foreground}

\pgfmathdeclarefunction{gauss}{2}{%
	\pgfmathparse{1/(#2*sqrt(2*pi))*exp(-((x-#1)^2)/(2*#2^2))}%
}


\tikzstyle{loss}=[draw, fill=none, text width=10em,
      text centered, minimum height=2.5em]

\tikzstyle{data}=[draw, fill=none, text width=10em, 
    text height=3.5cm, text centered, minimum height=12em, dotted]

\tikzstyle{ResDetail}=[draw, fill=yellow!10, text width=16em, 
text height=3.5cm, text centered, minimum height=12em, ]   

\tikzstyle{resblockA} = [fill=yellow!10, draw, text centered,  text width=7em,  
    minimum height=5em, rounded corners]
\tikzstyle{resblock} = [fill=white, draw, text centered,  text width=7em,  
minimum height=3em, rounded corners]

\tikzstyle{oper} = [fill=white, draw, text centered,  text width=10em,  
minimum height=1.3em, rounded corners]
\tikzstyle{sumoper} = [circle,fill=white, draw, text centered,  text width=0.7em,  minimum height=0.7em]

\tikzstyle{thickline} = [dotted, line width=0.5mm, draw = black]

\def\blockdist{3}
\def\edgedist{2.3}
\def\convdist{0.4}

\tikzset{global scale/.style={
		scale=#1,
		every node/.style={scale=#1}
	}
}
\begin{tikzpicture}

\node (rn6) [resblockA] {6 $\times$ ResNet Blocks};
    
\path (rn6)+(-\blockdist-1.0,0) node (data) [data] {Input Data};
\draw [->]  (data.east) to [] (rn6.west);
	
\path (rn6.50)+(1.2*\blockdist,0) node (resblock1) [resblock, draw=blue] {$Conv(3\times 3, 2)$ (\textit{Semantic})};
\draw [->, draw=blue]  (rn6.east) to [out=-10,in=180] (resblock1.west);

\path (rn6.-50)+(1.2*\blockdist,0) node (resblock2) [resblock, draw=black!30!green] {$Conv(3\times 3, 1)$ (\textit{Density})};
\draw [->, draw=black!30!green]  (rn6.east) to [out=-10,in=180] (resblock2.west);

\path (rn6)+(3.0*\blockdist,0) node (loss) [loss] {$ \textcolor{blue}{L_{semantic}}+\textcolor{black!30!green}{L_{density}}$};
\draw [->, draw=blue]  (resblock1.east) to [out=0,in=180] (loss.west);
\draw [->, draw=black!30!green]  (resblock2.east) to [out=0,in=180] (loss.west);

\path (loss)+(-5,-4) node (densgt) [data,draw=blue] { ${y}_{density} $};
\path (loss)+(-1,-4) node (labelgt) [data,draw=black!30!green] { ${y}_{label} $};
\draw [draw = blue]  (densgt.north) to [out=10,in=210] (loss.west);
\draw [draw = black!30!green]  (labelgt.north) to [out=110,in=210] (loss.west);


\path (rn6)+(-1.5,-4.1) node (resdetail) [ResDetail] {};
\node (resdetailname) [left=0.4cm of resdetail,] {\rotatebox{90}{ResNet Block Detail}};

\draw [loosely dashed]  (rn6.south) to [out=270, in=90] (resdetail.north);

\begin{pgfonlayer}{foreground}

	\path (resdetail.north)+(0,-\convdist-0.2) node (oper1) [oper] {$Conv(1\times1; 64) + BN$};
	\path (oper1.south)+(0,-\convdist) node (oper2) [oper] {$Conv(3\times3; 64) + BN$};
	\path (oper2.south)+(0,-\convdist) node (oper3) [oper] {$Conv(1\times1; 256) + BN$};
	\path (oper3.south)+(0,-\convdist) node (oper4) [sumoper] {$+$};
	\path (oper4.south)+(0,-\convdist) node (oper5) [oper] {ReLU};
	
	\draw [->]  (resdetail.north) to  (oper1.north);
	\draw [->]  (oper1.south) to  (oper2.north);
	\draw [->]  (oper2.south) to  (oper3.north);
	\draw [->]  (oper3.south) to  (oper4.north);
	\draw [->]  (oper4.south) to  (oper5.north);
	\draw [->]  (oper5.south) to  (resdetail.south);
	
	\path (oper1.north)+(0,0.2) coordinate (oper1upper);
	\path (oper2.west)+(-0.7,0) coordinate (oper2left);
	
	\draw [-]  (oper1upper) to [out=180,in=90] (oper2left);
	\draw [->]  (oper2left) to [out=270,in=170] (oper4.west);

\end{pgfonlayer}

\path (data)+(0,0.3) node (a) {\includegraphics[width=8em]{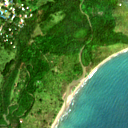}};
\path (densgt)+(0,0.3) node (a1) {\includegraphics[width=8em]{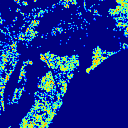}};
\path (labelgt)+(0,0.3) node (a2) {\includegraphics[width=8em]{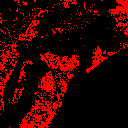}};




\end{tikzpicture}

%% file: experiments.tex
\section{Experiments}\label{sec:experiments}

In order to test how robust deep semantic density estimation is to changes in texture, object density and size, we create four different datasets. Three datasets contain trees (olive, coconut, oil palm) with different planting patterns and one dataset contains cars\footnote{
	Cars are reacquired VW Diesels sitting in a desert graveyard at the Southern California Logistics Airport in Victorville, USA.}.
A summary of all datasets is given in \Cref{tab:objects}. Note that object sizes range between $1/3$ and $1/4$ of a Sentinel-2 pixel (Ratio Areas). Object densities of the tree datasets are around 1, whereas the car dataset has highest object density ($>5$ cars per pixel on average) and smallest object size.
Our semi-automated groundtruth annotation procedure involving Faster R-CNN allows us to massively extend our label set across large regions to $>200 km^{2}$ with $>1.6$ Mio. object instances in total (Tab.~\ref{tab:objects}). We validate outcomes of Faster R-CNN predictions on small, manually labeled hold out regions. Intersection over Union (\textit{IoU}) ranges between $0.82$ and $0.97$ for Coconut and Olives, respectively. We achieve highest recall for cars ($0.92$) and lowest for coconut ($0.77$). 
%
\newcommand*\rfrac[2]{{}^{#1}\!/_{#2}}
\begin{table}
\centering
\begin{tabular}{p{2cm} S[table-auto-round, table-format=1.1, detect-weight] S[table-auto-round, table-format=3.1e1, detect-weight] *{2}{S[table-auto-round, table-format=1.2, detect-weight] } } 	\toprule 
 \multirow{2}{*}{ Dataset } & { Area } & { Object } & { Object }  & { Ratio } \\ 
 & { ($km^{2}$) } & {Count} & {Density\textsuperscript{a}}  & {Areas\textsuperscript{b}}  \\ \toprule
Coconuts &77.86277& 258.716765625 e3 & 0.966530873157	& 0.36\\
Palm & 104.62310  & 537.47475 e3 & 1.1610338005 & 0.36 \\
Olives & 128.96356& 853.743875 e3 & 0.904141794957 & 0.390625 \\
Cars & 0.81423 & 18.284 4023438 e3 & 5.35778671921 & 0.2625
\end{tabular}
\caption{Datasets: \textsuperscript{a}Mean object density per pixel excluding object free pixels. \textsuperscript{b}Ratio areas = $ {Area_{Object}}/{Area_{Pixel}} $}
\label{tab:objects}
\end{table}
%

%
%
%
\subsubsection{Semantic Segmentation and Density Estimation}

\begin{figure}[t!]
	\centering
	\begin{subfigure}[b]{0.49\textwidth}
		\includegraphics[width=0.49\textwidth, angle =90]{{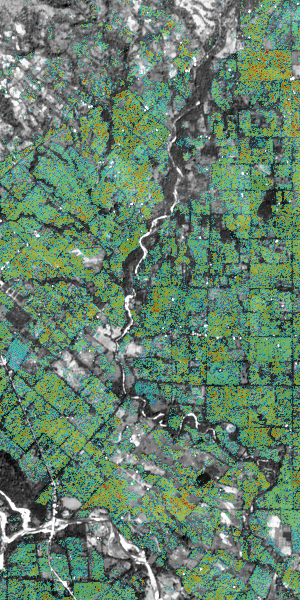}}
		\includegraphics[width=0.385\textwidth,  angle =90]{{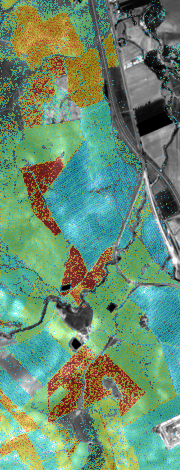}}
		\includegraphics[width=1\textwidth,trim=10cm 0cm 13cm 0cm, clip=true]{{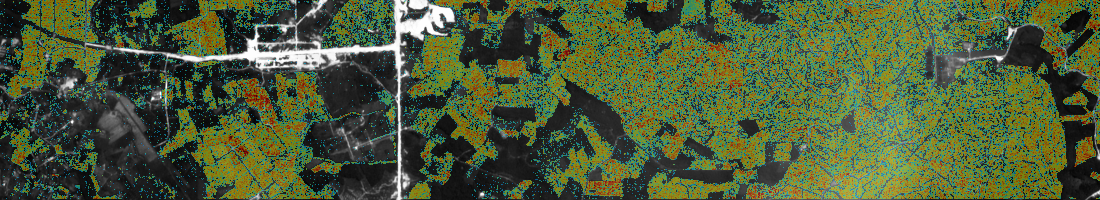}}
		\subcaption{GT}
	\end{subfigure}
	\begin{subfigure}[b]{0.49\textwidth}
		\includegraphics[width=0.49\textwidth,  angle =90]{{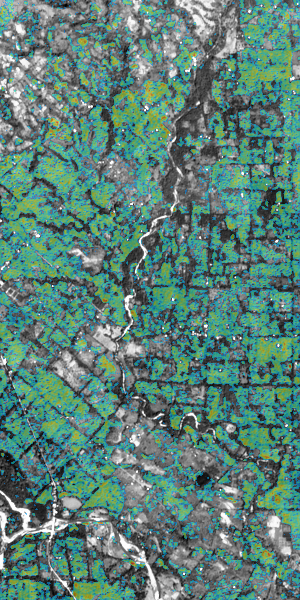}}
		\includegraphics[width=0.385\textwidth, , angle =90]{{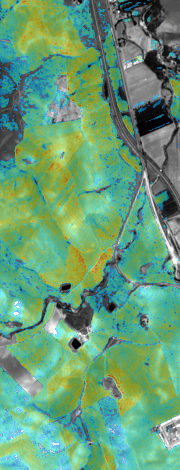}}
		\includegraphics[width=1\textwidth,trim=10cm 0cm 13cm 0cm, clip=true]{{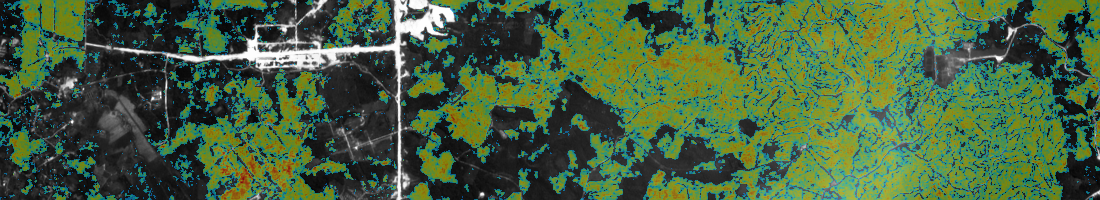}}
		\subcaption{Pred}
	\end{subfigure}
	\begin{subfigure}[b]{0.85\textwidth}
		\includegraphics[width=1\linewidth]{{figures/density_bar_0_0-2_0.png}}
	\end{subfigure}	
	\caption{ Density estimation results. Top: Coconut; Middle: Olives; Bottom: Palm. Densities below 0.5 were trimmed for visualization.}
	\label{fig:gt_pred_density}
\end{figure}

\begin{figure}[t!]
	\centering
	\begin{subfigure}[b]{0.49\textwidth}
		\includegraphics[width=0.57\textwidth, angle=90]{{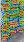}}
		\subcaption{GT}
	\end{subfigure}
	\begin{subfigure}[b]{0.49\textwidth}
		\includegraphics[width=0.57\textwidth, angle = 90]{{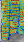}}
		\subcaption{Pred}
	\end{subfigure}
	\begin{subfigure}[b]{0.85\textwidth}
		\includegraphics[width=1\linewidth]{{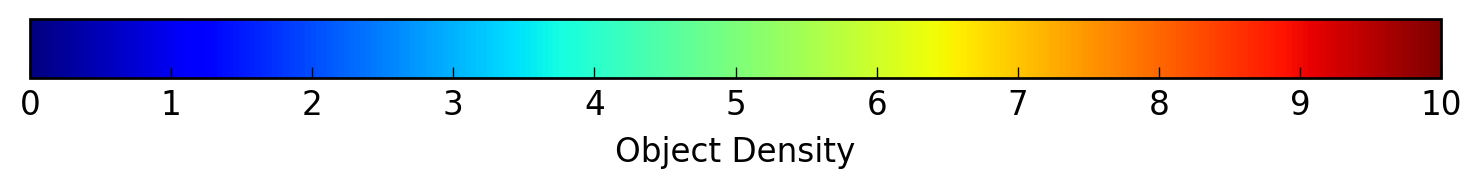}}
	\end{subfigure}	
	\caption{Car density, densities below 0.5 were trimmed for visualization.}
	\label{fig:gt_pred_density_cars}
\end{figure}

Results of deep semantic density estimation for all architectures are presented in \Cref{tab:res_sem_seg} and absolute numbers of object counts are shown in \Cref{tab:res_obj_counting}. A visual comparison of ground truth and predictions in Figs. \ref{fig:gt_pred_density} \& \ref{fig:gt_pred_density_cars} and the respective $\chi^2$ distance between ground truth distribution $y_{density}$ and predicted distribution $\hat{y}_{density}$ (\Cref{tab:res_sem_seg}, right column) show good performance. \emph{Ours} and \emph{Ours atrous} architectures outperform standard computer vision architectures that build in many downsampling (and upsampling) steps in their network. Results for cars show that there is a natural limitation of our method in terms of minimum object size and maximum object density. While cars can be identified very well (high \textit{IoU}), there exact number is much harder to estimate (higher \textit{MSE}, \textit{MAE} compared to the tree datasets). Moreover, \textit{MSE} is significantly higher than the \textit{MAE}, which is caused by outliers in high density areas where lower densities are erroneously predicted (Fig.~\ref{fig:gt_pred_density_cars}). Our method seems to have a slight tendency to underestimate high density areas that are surrounded by regions of much lower density. This shortcoming of our method can be observed if comparing ground truth high density areas of olives (red) with the respective, underestimated predictions in the middle row of \Cref{fig:gt_pred_density}. This effect translates to slightly underestimating the number of instances globally by $<5\%$ for all datasets (cf.Tab.~\ref{tab:res_obj_counting}).  
\begin{table}[h!]
	\centering
	\begin{tabular}{cc *{6}{S[table-auto-round, table-format=1.3, detect-weight] } S[table-auto-round, table-format=1.1e1, detect-weight] }
		\toprule
\multirow{2}{*}{Object} & \multirow{2}{*}{Architecture} & \multicolumn{3}{c}{Semantic Segmentation} &&  \multicolumn{3}{c}{Density}  \\\cmidrule{3-5} \cmidrule{7-9}
		 &  & IoU & {Precision} & {Recall}  && {MSE} & {MAE}& { $\chi^2$ Distance} \\ \toprule
\multirow{4}{*}{Coconut} & DL2 & 0.459009927026 & 0.673019457718 & 0.590750858193 && 0.191485866904 & 0.354863971472 & 69.6772774222 e3 \\
						& DL3 & 0.52788837784 & 0.645370428652 & 0.743581685442 && 0.16765126586 & 0.340611189604 & 62.4904302391 e3\\
& Ours & 0.624585834809 & \bfseries 0.774609015639 & 0.76330803075 && 0.126230448484 & \bfseries 0.264597713947 & \bfseries 50.55666649 e3 \\
& Ours Atrous & \bfseries 0.64871731199 & 0.755595603613 & \bfseries 0.820988251221 && \bfseries 0.122092694044 &  0.27280947566 & 81.6085640588 e3
\\ \cmidrule{2-9}
\multirow{4}{*}{Palm} & DL2 & 0.564694276264 & 0.709796897541 & 0.734205761041 && 0.306953430176 & 0.42098647356 & 93.7745183001 e3\\
& DL3 &  0.579727241519 & 0.694115624907 & 0.778654304014 && 0.276303201914 & 0.415431022644 & 88.596 5882869 e3\\
& Ours & \bfseries 0.660110269834 &\bfseries 0.7945818827 & 0.795940779987 && \bfseries 0.191842526197 & \bfseries 0.313708215952 & \bfseries 75.8253116923 e3\\
& Ours Atrous &  0.542233147482 & 0.544303681137 & \bfseries 0.993033432816 && 0.671634018421 & 0.668232500553 & 97.1177808153 e3 \\ \cmidrule{2-9}
\multirow{4}{*}{Olives} & DL2 & 0.777372376102 &\bfseries 0.920622954437 & 0.833219616914 && 0.234816223383 & 0.362933069468 & \bfseries 30.8545857805 e3\\
& DL3 &0.811314871186 & 0.882795837911 & 0.909254297649 && 0.174634695053 & 0.331230551004 & 35.587368922 e3\\
& Ours & \bfseries 0.861348713933 & 0.900598330972 & 0.951839709774 && \bfseries 0.135416164994 &\bfseries 0.270482957363 & 34.471 6724155 e3 \\
& Ours Atrous & 0.858089552239 & 0.892719057157 & \bfseries 0.956748930788 && 0.161898553371 & 0.302339732647 & 46.4968425521 e3 \\ \cmidrule{2-9}
\multirow{4}{*}{Cars}	& DL2 &0.789221556886 & 0.928169014085 & 0.84056122449 && 8.74258708954 & 2.23502063751 & 21.310907774 e1\\
& DL3 &0.753186558517 & 0.891632373114 & 0.829081632653 && 8.45434570312 & 2.31883215904 & 25.0679757085 e1\\
& Ours &0.930704898447 & 0.936298076923 & 0.99362244898 && 2.31908679008 & 1.0769701004 &\bfseries 1.5720852052 e1 \\
& Ours Atrous & \bfseries 0.940963855422 & \bfseries 0.944377267231 & \bfseries 0.996173469388 && \bfseries 1.94000828266 & \bfseries 0.991350054741 & 3.75608152846 e1\\
	\end{tabular}
	\caption{Semantic Segmentation and Density Estimation over test areas. In \textbf{bold} the best performing metric per object class. $\chi^2$ Distance is the histogram distance between ground truth $y_{density}$ and predicted $\hat{y}_{density}$}
	\label{tab:res_sem_seg}
\end{table}
\begin{table}
	\centering
	\begin{tabular}{c *{2}{S[table-auto-round, table-format=3.2e1, detect-weight]} S[table-auto-round, table-format=2.2, detect-weight]}
		\toprule
		\multirow{2}{*}{Object} & \multicolumn{3}{c}{Object Count}  \\\cmidrule{2-4} 
		&   {Ground Truth} & {Prediction} & { Diff \%}   \\ \toprule
		Coconut &    88.4905 e3& 84.6475703125 e3& -004.34275865555 	  \\
		Palm  &   143.13571875 e3& 137.127453125 e3& -004.19760346413 		  \\
		Olives  &   64.0528203125 e3 & 62.338390625 e3 & -002.67658829689  \\
		Cars & 4.32876074219 e3 & 4.13069335938 e3 & -004.57561612129 \\
	\end{tabular}
	\caption{Object Counting Performance with lowest MSE and MAE error. \emph{Ours} for tree objects and \emph{Ours Atrous} for Cars.}
	\label{tab:res_obj_counting}
\end{table}

\subsubsection{Comparison of architectures}

\begin{figure}[t!]
	\centering
 \begin{subfigure}[b]{0.49\textwidth}
	\includegraphics[width=1\linewidth]{{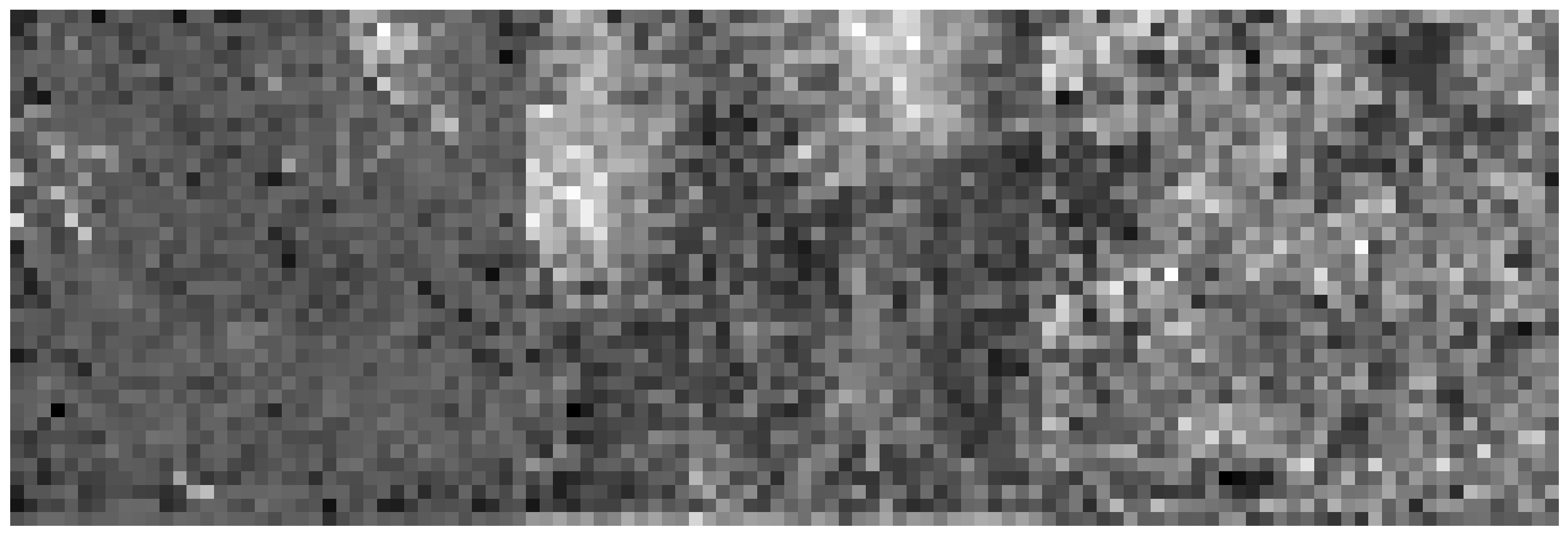}}
\subcaption{DL2}
 \end{subfigure}
 \begin{subfigure}[b]{0.49\textwidth}
 	\includegraphics[width=1\linewidth]{{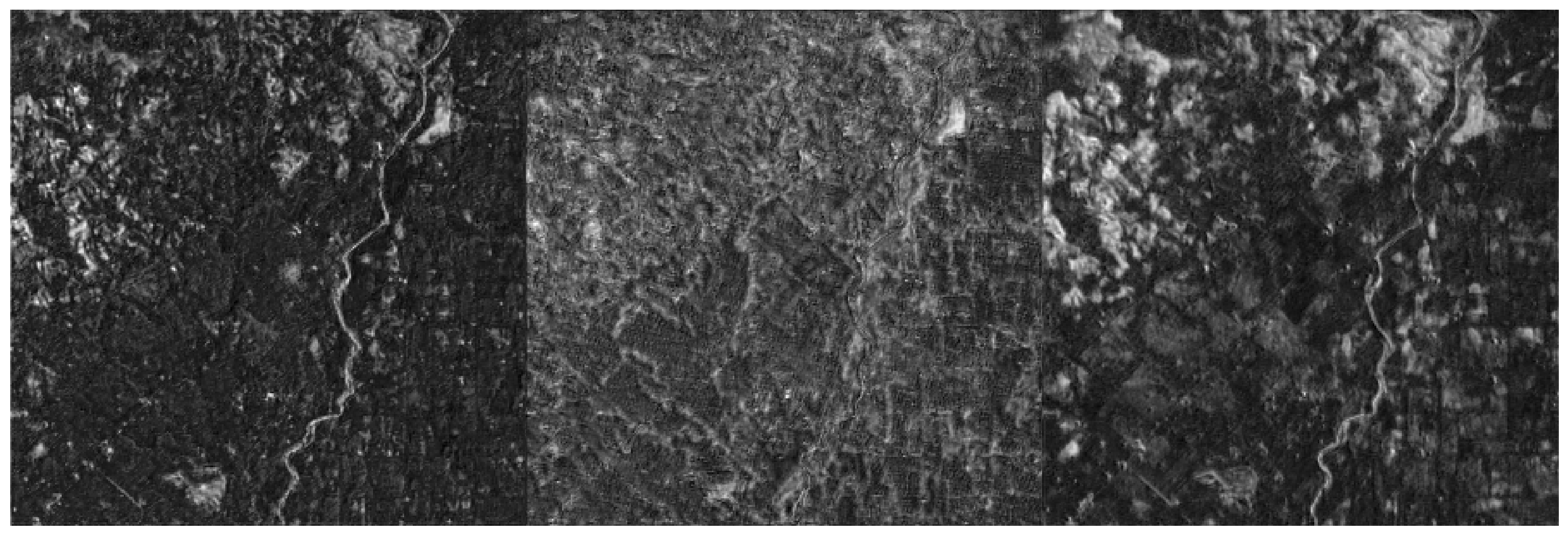}}
 	\subcaption{Ours}
 \end{subfigure}
	\caption{Final activation maps of (a) \emph{DL2} and (b) \emph{Ours} for a test region of the coconut dataset. \emph{Ours} better retains details than \emph{DL2}.}
	\label{fig:activations}
\end{figure}

Our simple architectures \emph{Ours} and \emph{Ours Atrous} consistently outperform \emph{DL2} and \emph{DL3}. \emph{DL2} and \emph{DL3} use strided convolutions to learn lower dimensional representations of the image, which generates higher level features over large areas to learn context. This seems less important in cases where the extent of individual objects in the image is small compared to the pixel size. In fact, strided convolutions and forcing the network to learn lower dimensional representations risks loosing high frequency information in the satellite images. 
We compare activation maps of architectures \emph{DL2} and \emph{Ours} on the coconut dataset in Fig.~\ref{fig:activations} to visualize this argument. One can observe that filters learned by \emph{DL2} loose details (Fig.~\ref{fig:activations}(a)) required for a fine-grained density prediction in contrast to \emph{Ours} (Fig.~\ref{fig:activations}(b)). Recall that both, \emph{DL2} and \emph{DL3}, use bicubic upsampling at different rates to obtain predictions at the original scale. By removing strided convolutions in our architectures, we avoid the need for upscaling predictions already from the start. 


\subsubsection{Band importance}

Sentinel-2 satellites are multi-spectral sensors designed for vegetation monitoring. Spectral signatures of different vegetation types and plant species can help distinguishing objects, in contrast to mostly texture-based features in RGB images. In order to verify the contribution of this additional spectral evidence, we train and test \emph{Ours Atrous} with different band settings. Each run leaves away bands to test their importance. We contrast results on vegetation with results for cars. We show results of this analysis in Tab.~\ref{tab:band_importance}, where high \textit{IoU} values of column \textit{Semantic} indicate good identification of objects while low \textit{MSE} and \textit{MAE} values of \textit{Density} mean correct object counting. It turns out that deep semantic density estimation requires both, spectral and spatial evidence for vegetation whereas this is not the case for cars. 10m RGB bands help counting trees (low \textit{MSE} and \textit{MAE} with RGB bands) whereas infared bands help identifying them (high \textit{IoU} for \textit{All} and \textit{No RGB}). 
%
Infrared bands are less important for cars because, unlike plants, they do not reflect characteristic infrared signatures. Prediction is based mostly on texture in the image provided by the 10m RGB bands. 

\begin{table}[h!]
	\centering
	\begin{tabular}{cc *{10}{S[table-auto-round, table-format=0.2, detect-weight]}}
		\toprule
\multirow{3}{*}{Bands}	& \multirow{3}{*}{\shortstack{Original Pixel \\Resolution $[m]$} } && \multicolumn{4}{c}{Coconut} && \multicolumn{4}{c}{Cars} \\ \cmidrule{4-7} \cmidrule{9-12}
&&& {Semantic} && \multicolumn{2}{c}{Density} && {Semantic} && \multicolumn{2}{c}{Density} \\ \cmidrule{4-4} \cmidrule{6-7} \cmidrule{9-9} \cmidrule{11-12}
&   && {IoU} && {MSE} & {MAE} && {IoU} && {MSE} & {MAE}\\ \toprule 
  All &10 \& 20 && \bfseries 0.660429596901 &  & \bfseries 0.127538889647 &\bfseries  0.276313453913 &&\bfseries 0.821963012218 &  & \bfseries 3.55680370331 & \bfseries 1.39161252975 \\
  RGB  &10 &&  0.493429660797 &  & 0.26147967577 & 0.40706294775 && \bfseries0.829918980598 &  &\bfseries 3.52916526794 & \bfseries 1.39942765236\\
  RGB I & 10 &&  0.550070285797 &  &\bfseries 0.161799028516 &\bfseries 0.325526088476 & & 0.766188502312 &  & 3.67557024956 & 1.41544449329\\
  No RGB & 10 \& 20 &&\bfseries 0.627096951008 &  & 0.194284856319 & 0.347905695438 &&0.806366562843 &  & 4.45285558701 & 1.56986606121\\
	\end{tabular}
	\caption{Band Importance: \textbf{top two} per dataset. All: 10m and 20m bands. RGB: B2, B3, B4. RGBI: RGB plus infrared (B8). No RGB: all 10m and 20m bands except RGB.}
	\label{tab:band_importance}
\end{table}

%% file: conclusion.tex
%
%
\section{Conclusion}

We have proposed end-to-end learnable deep semantic density estimation for counting object instances of fine-grained classes in cluttered background. Results show that counting objects of sub-pixel size is possible for 10m GSD satellite images. Experimental evaluation with four datasets shows that our method is robust to change in object types, background, and object density. It turns out that a shallow network specifically designed for satellite imagery of 10m GSD and sub-pixel objects outperforms more sophisticated, state-of-the-art architectures from computer vision. This signifies that direct application of networks tailored for vision to remote sensing images should be done with care. In our experiments, we find that any down-sampling operation inside the network risks loosing precious details. We should thus always keep in mind the particularities of remote sensing imagery in terms of object scale, GSD, (nadir) perspective and (high) spectral resolution. If carefully considered during the network design process, these specific properties offer new possibilities in network design. 